# Vision System and Depth Processing for DRC-HUBO+

Inwook Shim[1], Seunghak Shin[1], Yunsu Bok[1], Kyungdon Joo[1], Dong-Geol Choi[1],
Joon-Young Lee[1], Jaesik Park[1], Jun-Ho Oh[2], and In So Kweon[3]

*Abstract*— This paper presents a vision system and a depth processing algorithm for *DRC-HUBO+*, the winner of the *DRC finals* 2015. Our system is designed to reliably capture 3D information of a scene and objects robust to challenging environment conditions. We also propose a depth-map upsampling method that produces an outliers-free depth map by explicitly handling depth outliers. Our system is suitable for an interactive robot with real-world that requires accurate object detection and pose estimation. We evaluate our depth processing algorithm over state-of-the-art algorithms on several synthetic and real-world datasets.

## I. INTRODUCTION

The necessity of a substitute robot for person has come to the fore since the Fukushima Daiichi nuclear disaster on March 11, 2011. For counteracting and assisting humans in responding to the disaster, the US Defense Advanced Research Project Agency (*DARPA*) held the *DARPA* Robotics Challenge (*DRC*) in 2013 (*trials*) and 2015 (*finals*) [1]. In the challenges, a robot should carry out diverse tasks with a limited human-robot interaction, therefore recognizing surrounding environment and objects becomes one of the fundamental abilities of the robot.

In recent research, depth sensors are widely used in the computer vision and robotics field, and open a new horizon for scene understanding [2] and object recognition [3] since they give rich information of a scene in real time. There are various depth sensors such as stereo-based range sensors, 3D time-of-flight (*3D-ToF*), active pattern cameras (*e.g.Microsift Kinect*), and Light Detection and Ranging (*LiDAR*). Among them, a lidar sensor has a wide measurable range and is also robust to the effects of sunlight, therefore it is considered as the most suitable sensor for outdoor robots.

Although a lidar sensor is quite reliable, the depth data from a lidar sensor has a form of sparse point clouds, which is typically less than the resolution of an image sensor. In addition, the measured depth may contain depth noise and *flying points* around depth boundaries. This issues make recognizing objects and estimating their poses more challenging while this is one of core techniques for robotics applications. In many cases, the failure of object pose estimation may cause a fatal accident.

To handle these problems, we follow a common principle of depth upsampling which propagates sparse depth points by utilizing sharp edge boundaries of the corresponding image. Many studies have been conducted to achieve an accurate and dense depth map from sparse observed points. However, these techniques usually assume that a depth map and image pair is well-aligned and has ignorable alignment error. This assumption is not appropriate especially a depth and camera pair has a wide baseline which is common in robotics applications because the wide baseline sensors generate a large parallax effect to captured data. As a result, the projected depth points in an image domain exhibit *flipping points* and *depth dis-occlusion*.

In this paper, we present a complete system composed of vision sensors and a depth processing algorithm, and show its applications in the mission of the *DRC finals* 2015. We design our sensor system with a pair of a color camera and a lidar. We obtain the 3D structure data of a target area aligned with image information by rotating the sensor. We also propose a new method to obtain a high-resolution depth map by explicitly considering the alignment problem between image and depth data. The key to our upsampling technique includes handling flying points, flipping points, and dis-occlusion region and estimating a confidence map to remove unreliable data. The proposed depth upsampling method was evaluated on benchmark datasets. We also demonstrate the real-world robotics applications such as 3D object pose estimation and toehold detection which are used for *DRC finals*. Our sensor system and the propose algorithm were adopted for *DRC-Hubo+* which was declared the winner of the *DRC finals* 2015.

## II. RELATED WORK

We review relevant hardware systems and depth processing algorithms that exploit depth sensor and color camera.

**Sensor system** Two representative and unveiled sensor systems in the *DRC* used for obtaining visual and geometric information of a scene are: *Team ihmc robotics* [4] and *Tartan Rescue* [5]. *Team ihmc robotics* uses Multisense-SL designed by *Carnegie Robotics* [6]. This sensor system consists of a forward-facing stereo camera and an rotating axial lidar, and it additionally has two wide-angle cameras to give an operator visual monitoring. *Tartan Rescue* designs their own sensor system which consists of two lidars, two stereo cameras, and two fisheye cameras.

**Depth upsampling** Given a depth and color image pair, depth upsampling approaches output a high-quality dense depth map that follows crisp edge of the color images. Joint bilateral upsampling (JBU) [7] applies spatially varying filter to the sparse samples. The filtering kernels are determined by local color affinity and radial distance. Chan *et al*. [8] accelerate the JBU using a GPU and introduce depth noise aware kernel. Dolson *et al*. [9] present a flexible high-dimensional filtering method for increased spatial and temporal depth resolution. Park *et al*. [10], [11] use a least-square cost function that combines several weighting factors together with nonlocal structure. Ferstl *et al*. [12] design smoothness term as a second order total generalized variation, and propagate sparse seed points using anisotropic diffusion tensor obtained from a high-resolution image.

[1]I. Shim, S. Shin, Y. Bok, K. Joo, D.-G. Choi, J.-Y. Lee, and J. Park are with the the Robotics and Computer Vision Laboratory, Department of Electrical Engineering, KAIST, Daejeon, 305-701, Korea {iwshim, shshin, ysbok, kdjoo, dgchoi, jylee, jspark}@rcv.kaist.ac.kr

[2]J.-H. Oh is with Humanoid Research Center, School of Mechanical, Aerospace and Systems Engineering Department of Mechanical Engineering, KAIST, Daejeon, 305-701, Korea jhoh@kaist.ac.kr

[3]I.S. Kweon is with Department of Electrical Engineering, KAIST, Daejeon, 307-701, Korea iskweon@kaist.ac.kr

Among them, filtering approaches [7], [8], [9] have low computational complexity and can be easily parallelized for real-time applications. However, it may not cover large depth holes which indicates physically unmeasurable depth samples occurred by disocclusion or distant scene such as sky. In contrast, the global approaches [10], [11], [12] can fill out the large depth holes because the entire depth variables in the cost function are densely connected. However it may also propagate erroneous observations to the large area.

**Depth outlier handling** In practice, sparse seed points used for depth upsampling may contain outliers. This is because of *flipping points* and *depth dis-occlusion* occurred when the measured range data is projected onto image domain. In addition, there are *flying points*, which indicate intermediate depth value between foreground and background depth occurred around object boundaries. To overcome this challenges, Kang *et al*. [13] detect the flipping points based on the distribution of depth values within a color image segment. Park *et al*. [10] measure depth variation of the local regions, which is for detecting depth discontinuity. Those discontinuity candidate regions are refined via binary label optimization. The extended work [11] proposed heuristic approach that detects flipped depth orders after depth projection. However, their work evaluates the performance of algorithm on the exactly-aligned depth-color image pairs. In the wider extent, ToF depth camera and stereo color camera fusion [14], [15] are also introduced. Gandhi *et al*. [15] look for small region that have mixed foreground and background depth samples. Georgios *et al*. [14] grow the seeds using a smoothness term that is conditionally defined by occlusion label obtained from depth discontinuity analysis.

In this paper, we focus practical issues on an image and depth capturing system and depth processing approach for robot applications. We introduce sensor calibration for our vision system, depth outliers handling, and depth map upsampling. Our filter based approach can generate a high-quality and outliers-free depth map that is suitable for accurate object detection and pose estimation. Our depth outliers rejection stage does not dependent on the edge information of the image, therefore it keeps reliable points even in ambiguous image edges. In addition, our confidence map explicitly disregards large holes in a depth map and propagates reliable samples.

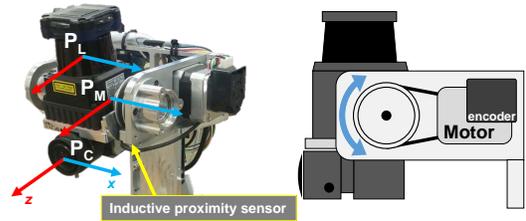

Fig. 1: Sensor system configuration (left) and an example of its movement (right). lidar and camera are placed on the same plate, which is rotated by a single motor.

## III. SENSOR SYSTEM

### A. System Configuration

Fig. 1 shows our sensor system which consists of one lidar sensor, one *GigE* camera, and one stepper motor with an incremental encoder. The camera equipped with a 2.2*mm* lens has 1288×964 resolution with a 118.6°×90.0° field-of-view (FoV). The motor with a high resolution encoder provides angle information in 0.02 degree resolution. The motor finds its zero reference position by checking the signal of the inductive proximity sensor.

This system acquires 3D points by rotating the lidar sensor around the motor's *x*-axis and captures an image at each target angle. We can control the rotation scope, rotation speed, and target angle using our control software, therefore we can control the sparsity of 3D points by trading off capturing time when we reconstruct the 3D structure of a target area.

### B. System Calibration

The sensor system should be calibrated in the robot coordinate so that captured data can be utilized for robot applications. We divide the whole process of the calibration into three steps; (1) camera intrinsic, (2) camera and motor, and (3) camera and lidar extrinsic.

The first step, camera intrinsic calibration, is done by a conventional method using a planar checkerboard pattern [16]. To account for the short focal length of the lens, the conventional model adopted a fisheye distortion model [17] as follows:

$$\begin{bmatrix} x_d \\ y_d \end{bmatrix} = \frac{\tan r}{r} \begin{bmatrix} x \\ y \end{bmatrix}, \quad (1)$$

$$r = \sqrt{x^2 + y^2}, \quad (2)$$

where $[x \ y]^\top$ and $[x_d \ y_d]^\top$ denote the ray directions in the camera coordinate system ($z = 1$) before and after distortion, respectively. The model is derived from the equidistant model among the mapping functions of fisheye lenses [17]. However, the model could not reduce the projection error sufficiently. We modify the distortion model of the fisheye lens by adding an additional parameter *k* to *r*.

$$r' = k\sqrt{x^2 + y^2}, \quad (3)$$

where *k* is an unknown distortion parameter. Substituting Eq. (2) with Eq. (3), the mean projection error is reduced from 1.692 pixels to 0.360 pixels, where the value of *k* is estimated as 0.9413.

For the next two steps, we designed a pattern that consists of two perpendicular checkerboard patterns, rather than using a single pattern. Scan data on both planes are extracted and used to estimate the relative pose between a camera and a lidar sensor as follows.

For each pose of the pattern, we capture images for every 10 degrees of motor rotation. Although we use a high-accuracy step motor, we assume that only its repeatability (not its angular accuracy) is reliable. The rotation angles of the images are considered as unknown values, rather than fixed-value constraints. Let $A_\theta$ be the unknown angle corresponding to the motor angle $\theta$ computed by counting motor steps. In our implementation, the angle $\theta$ varies from $-30$ to $80$ degrees so that we add 11 unknown variables $A_{-30} \sim A_{80}$ corresponding to 12 angles. It should be noted that $A_0$ is fixed to zero as reference angle. Because we set the rotation axis of the motor as the x-axis of the motor coordinate system, the rotation matrix $\mathbf{R}_\theta$ is compute as follows:

$$\mathbf{R}_\theta = \begin{bmatrix} 1 & 0 & 0 & 0 \\ 0 & \cos A_\theta & \sin A_\theta & 0 \\ 0 & -\sin A_\theta & \cos A_\theta & 0 \\ 0 & 0 & 0 & 1 \end{bmatrix}. \quad (4)$$

We capture the images of the pattern in *N* different poses. The number of images in each pose may be different due to the limitation of the camera's FoV. Let $K_n$ be the number of frames captured in the *n*-th pose. The cost function $f_c$ for the estimation of the motor-to-camera transformation $\mathbf{H}_{mc}$ is the projection error of feature points on the perpendicular patterns:

$$f_c(\mathbf{H}_{mc}, \mathbf{H}_1 \sim \mathbf{H}_N, A_{-30} \sim A_{80})$$
$$= \sum_{n=1}^{N} \sum_{k=1}^{K_n} \sum_i \|\mathbf{q}_i - proj(\mathbf{H}_{mc}\mathbf{R}_\theta \mathbf{H}_n \mathbf{p}_i)\|^2, \quad (5)$$

where $\mathbf{H}_n$ denotes the transformation from the pattern in the *n*-th pose to motor and $proj(\cdot)$ indicates the process of the projection from camera to image, including radial distortion. $\mathbf{p}_i$ and $\mathbf{q}_i$ represent the feature points of the patterns and its location in the image, respectively. Because the pattern-to-motor and motor-to-camera transformations compensate each other, we fix both rotation and translation of the motor-to-camera transformation along x-axis as zero.

The last step, camera-lidar calibration, is easily done by utilizing the results of the previous step. Because we have already estimated the pattern-to-camera transformation for every image, we simply adopt the constraint that lidar scanning points must lie on the corresponding patterns [18]. The cost function $f_l$ of the estimation of the lidar-to-camera transformation $\mathbf{H}_{lc}$ is the distance between the lidar scanning points and planar patterns:

$$f_l(\mathbf{H}_{lc}) = \sum \left(\mathbf{v}_z^\top \mathbf{H}_{pc}^{-1} \mathbf{H}_{lc} \hat{\mathbf{p}}\right)^2, \quad (6)$$

where $\hat{\mathbf{p}}$ indicates the lidar scanning points on the patterns. We adopt the z-axis unit vector $\mathbf{v}_z = [0\ 0\ 1\ 0]^\top$ to consider only the z-terms of the points because we set each planar pattern as $z = 0$ of its own pattern coordinate system.

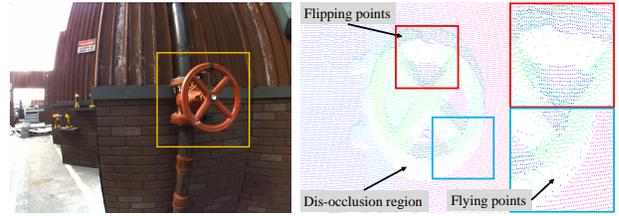

Fig. 2: The images show effects of the parallax effects. Left image shows an image of a target region, and right image depicts registered 3D points to the camera coordinate, $P_C$.

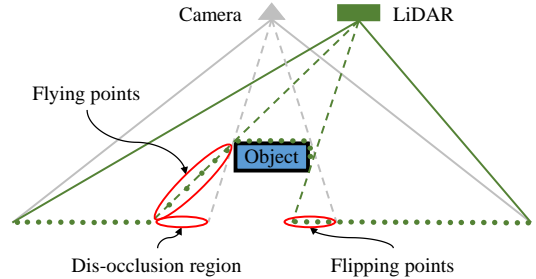

Fig. 3: This example shows why flying and flipping points, and dis-occlusion problems are occurred in the depth and image alignment.

## IV. DEPTH PROCESSING

With our system, we can capture dense 3D information by setting motor rotation speed slow. However, it is important to capture *high quality* 3D information within a reasonable *time budget* for the *DRC finals* since we should carry out the tasks quickly with a limited human-robot interaction. For satisfying the conflicting requirements, we adjust the motor rotation speed to be suitable for human-robot interactions and perform depth upsampling to complement sparse raw depth measurements using a high-resolution image.

The first step for depth upsampling is to align a depth and image pair. We align them by projecting depth measurements into the image domain using the calibration parameters estimated in the previous section. We show an example of the alignment in Fig. 2. As shown in the figure, we have observed that there are several erroneous depth points due to the several reasons depicted in Fig. 3.

The flying points are caused by measurement noise of depth sensor around object boundaries. The flipping points are occurred when the camera cannot see the corresponding region due to occlusion. Though they are regular points in the lidar coordinate, they are projected onto the occluding object region in the image coordinate. The dis-occlusion region is the opposite case of the flipping points. There is no true depth measurement due to occlusion. These alignment problems are amplified due to system calibration error and measurement noise of depth sensor. In all the image-guided depth upsampling algorithms, the unreliability of the aligned measurements degrades the performance of depth upsampling severely as will be shown in the Experiment Section V.

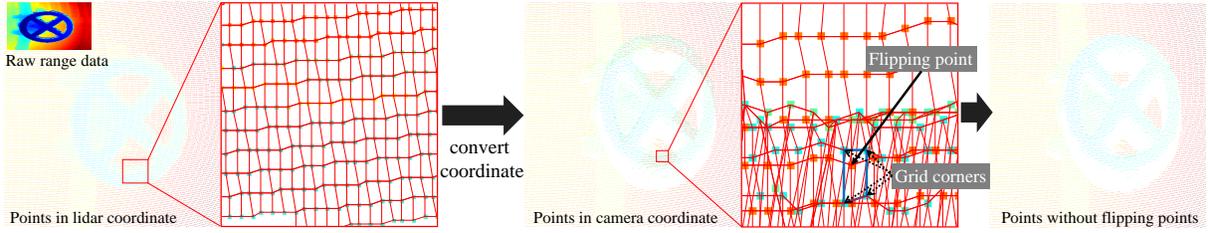

Fig. 4: Pipeline for eliminating flipping points using the 4-connected grid map. The flying and sparse points are removed in advance using the method described in Section IV-A.

Therefore, we have to handle the unreliability explicitly before performing depth upsampling. We remove the flying points if two adjacent points along a lidar scan-line have large distance, then apply a 2D filter to eliminate isolated sparse points in the image domain. We also remove the flipping points by checking the depth information among nearby points. After removing the suspicious points, we run our depth upsampling and generate a confidence map concurrently. Then, we use the confidence map to distinguish low reliability regions including the dis-occlusion. We describe more details in the following sections.

### A. Flying points rejection

The flying points are commonly occurred around object boundaries, since lidar measurements are noisy when a surface geometry is unstable. To remove flying points, we use a simple 1D filter as follows:

$$P_f = \{x | max(d(x_t^l, x_{t-1}^l), d(x_t^l, x_{t+1}^l)) > T_f\}, \quad (7)$$

where $P_f$ is a set of flying points, $d(\cdot)$ is an Euclidean distance between two points, $x_t^l$ is the $t$-th point in a $l$-th scan-line. $T_f$ is a predefined threshold and was empirically set to 30*mm* in our experiment. This filter is applied to each lidar scan-line. After that, we use morphological operations in the image domain for removing isolated sparse points.

### B. Flipping points rejection

Most of depth upsampling methods assume that the pair of a high-resolution image and a low-resolution depth map is well aligned, and they do not treat the flipping points problem seriously. In real environments, the flipping points yield a serious problem in depth upsampling, therefore it should be accounted for.

Fig. 4 shows the process for eliminating flipping points. We first generate a 4-connected grid map from depth measurements in the lidar coordinate. Each cell in the grid map is composed of four corner points. Then, we move the grid map to the camera coordinate, and find the points that invade into an another grid cell as shown in the center image of Fig. 4. Among the points, we reject the point if its depth is distant than the depth of each corner point of the invaded grid cell. The depth map after rejecting the flipping points is shown in the rightmost image in Fig. 4.

### C. Depth Map Upsampling and Confidence Map Estimation

In this section, we describe our depth upsampling algorithm. We also present how to compute a confidence map and determine parameters.

*1) Depth Map Upsampling:* For robotics applications such as object detection and pose estimation, we upsample a captured depth map with the guidance of an aligned image. Our depth upsampling algorithm is based on a rolling guidance filter suggested by Zhang *et al.* [19]. The rolling guidance filter is an iterative joint filter method that can achieve scale-aware local operations, therefore it is especially useful for removing small-scale structures such as noise while performing edge-preserving upsampling.

In our upsampling algorithm, we extend the JBU [7] with an additional depth guidance term to prevent texture copying problem and use the extended JBU as a joint filter in the rolling guidance filter. Specifically, our upsampling algorithm is formulated as follows:

$$D_p^{t+1} = \frac{1}{N_p} \sum_{q \in \Omega(p)} exp(G_{p,q} + K_{p,q} + H_{p,q}) R_q,$$

$$\text{s.t.} \quad G_{p,q} = exp(-\|p-q\|^2 / 2\sigma_s^2),$$
$$K_{p,q} = exp(-\|I_p - I_q\|^2 / 2\sigma_i^2), \quad (8)$$
$$H_{p,q} = exp(-\|D_p^t - R_q\|^2 / 2\sigma_d^2),$$
$$N_p = \sum_{q \in \Omega(p)} exp(G_{p,q} + K_{p,q} + H_{p,q}),$$

where $I$, $R$, and $D^t$ denote a guidance image, an aligned sparse depth map, and an upsampled dense depth map after the $t$-th iteration, respectively. $p$ is a query point in the 2D image coordinate, and $\Omega(p)$ is a set of neighboring points from $R$ within a filter range. We use only sparse points $\Omega(p)$ from $R$ as supporting pixels for efficient computation. $\sigma_s$, $\sigma_i$, and $\sigma_d$ denote the standard deviations to control the influence each of the spatial similarity term $G$, the intensity similarity term $K$, and the depth similarity term $H$, on the filter weights. $N_p$ is a normalization factor of the weights. For an initial upsampled depth map $D^0$, we use the JBU [7] where $H$ is set to zero in Eq. (8).

The Eq. (8) iteratively enhances an estimated dense depth map, $D^t$. In the scheme, the depth guiding term $H$ has an important role. Using the aligned raw depth measurements $R$ where outliers are effectively rejected, $H$ suppresses error propagation and texture copying problems, which are often occurred in the JBU. Also, it gives key information in computing the confidence of an estimated depth map as will be described in the next section.

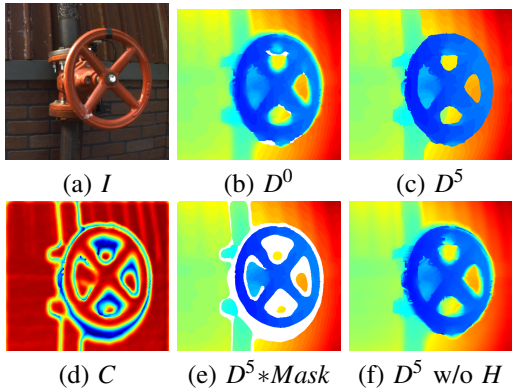

(a) $I$    (b) $D^0$    (c) $D^5$

(d) $C$    (e) $D^5 * Mask$    (f) $D^5$ w/o $H$

Fig. 5: Intermediate results of our upsampling method. (a) Input image, (b) Initial depth map $D^0$, (c) Our result after 5 iterations, (d) Confidence map, (e) Our result after masking low reliability regions in white. (f) Upsampling result without the $H$ term.

Fig. 5 shows the intermediate result of our upsampling method. In the figure, our result after five iterations in (c) has sharper and much accurate depth boundaries than an initial upsampled depth map in (b), while the result without the $H$ term in (f) has noisy depth boundaries due to the overfitting to intensity information.

*2) Confidence Map Estimation:* In our configuration that cannot ignore the baseline between depth and image sensors, there may exist ambiguity regions in upsampling results due to a lack of true measurements like dis-occlusion as shown in Fig. 3. While it is very difficult to solve the problem during upsampling process, we find that it can be effectively handled by computing a confidence map from the upsampling filter weights.

The confidence of depths is closely related to the statistics of measurements where small variance of local supporting measurements raises the confidence of resulting depth. Therefore, we define the confidence measure ($C$) of a up-sampled depth value on the location $p$ as:

$$C_p = \sum_{t=0}^{n} ( \sum_{q \in \Omega(p)} exp(G_{p,q} + H_{p,q})), \quad (9)$$

where $n$ is the number of iterations, and other notations are the same as Eq. (8). This confidence measure is simultaneously computed during processing our upsampling algorithm. The notion behind this measure is that a pixel has a low confidence if the estimated depth is supported from few or unstable depth measurements. From the measure, we mask an estimated depth out as a unreliable result if its confidence value is lower than $0.35 \times max(C)$.

Fig. 5-(d) shows our confidence map and (e) is the upsampling result with a confidence mask. As shown in the figure, our confidence mask effectively removes unreliable regions around depth boundaries due to parallax effects, and retains important depth information with clean and sharp depth boundaries.

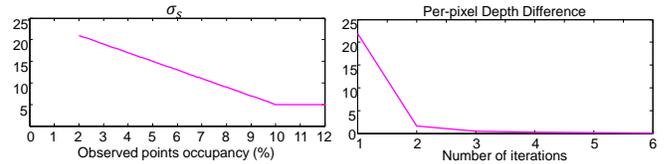

(a) Spatial smoothness parameter    (b) Number of iterations

Fig. 6: Parameter selection

*3) Parameter selection:* We have several parameters to perform our depth upsampling. $\sigma_s$ is a spatial smoothness parameter and we adaptively determined it through empirical cross-validation since the proportion of measured depth points to the guidance image pixels may vary according to the rotation speed of our system. Fig. 6-(a) shows the parameter we used according to the proportion. For example, if the measured points occupy 5% of a guided image area, $\sigma_s$ is set to 15. To guarantee the quality of our upsampling result, we control the maximum rotation speed of our sensor system to secure at least 2% of the proportion.

$\sigma_d$ is a depth similarity parameter to suppress depth measurement noise. We determined $\sigma_d$ according to the specification of our lidar sensor, *UTM 30LX-EW*. Because the maximum repeated accuracy of the lidar sensor is less than $\pm 30mm$, we set $\sigma_d$ to 30. We empirically set an intensity similarity parameter $\sigma_i$ to 20.

We should also determine the number of iterations in the rolling guidance scheme. We compute the average depth variations at each iteration step with an example dataset and depict the result in Fig. 6-(b). As stated in the original paper [19] of the rolling guidance filter, the depth map rapidly converges to the final result within 3~5 iterations.

## V. EXPERIMENTAL RESULTS

To validate the performance of the proposed algorithm including depth outliers rejection, depth upsampling, and confidence map, we perform experiments on both synthetic and real datasets, and compare our method with the state-of-the-art methods such as total generalized variation (TGV) [12], nonlocal means (Nonlocal) [10], and joint bilateral filter (JBU) [7]. For the experiments, we used two computers; a data capturing machine equipped with 1.3GHz dual core CPU and 8GB RAM, and a depth processing machine equipped with 3.6GHz quad core CPU and 16GB RAM. Our CPU-based implementation takes less than a second to generate a dense depth map of $640 \times 480$ resolution with five iterations of joint filtering.

### A. Middlebury Evaluation

For a quantitative evaluation, many existing papers assume that the pair of depth and image is exactly aligned [12], [10], [7]. In practice, there are many sources of alignment error including measurement noise, system calibration error, flying and flipping points, and dis-occlusion, therefore the aligned sparse depth samples on the image domain exhibit severe flipping as described in Fig. 2.

TABLE I: Quantitative comparison results. The detailed description is shown in Section V-A.

| Dataset | Adirondack | | Bicycle1 | | Classroom1 | | Flowers | | Motorcycle | | Storage | | Umbrella | | Vintage | |
|---|---|---|---|---|---|---|---|---|---|---|---|---|---|---|---|---|
| Error metric | A80 | A95 | A80 | A95 | A80 | A95 | A80 | A95 | A80 | A95 | A80 | A95 | A80 | A95 | A80 | A95 |
| TGV [12] | 19.6 | 152.3 | 14.5 | 86.8 | 40.2 | 364.3 | 64.5 | 1028.0 | 32.0 | 388.9 | 44.9 | 723.2 | 32.9 | 259.5 | 40.3 | 403.8 |
| Nonlocal [10] | 9.7 | 285.9 | 9.1 | 183.7 | 6.3 | 99.0 | 125.5 | 682.2 | 29.7 | 471.8 | 86.1 | 1084.8 | 8.2 | 229.4 | 8.9 | 84.5 |
| Bilateral [7] | 4.7 | 160.5 | 4.4 | 116.0 | 4.4 | 21.0 | 7.5 | 575.6 | 7.5 | 379.0 | 4.9 | 448.4 | 4.6 | 89.8 | 4.3 | 17.1 |
| Ours w/o $C$ | 4.0 | 8.4 | **3.6** | 8.0 | 3.6 | 9.0 | 3.7 | 7.6 | 5.7 | 15.5 | 3.9 | 10.4 | 3.6 | 7.4 | 4.6 | 8.1 |
| Ours | **3.3** | **7.0** | 4.5 | **6.4** | **3.2** | **6.3** | **3.3** | **5.7** | **5.0** | **9.9** | **3.6** | **7.9** | **3.5** | **6.4** | **4.4** | **7.5** |

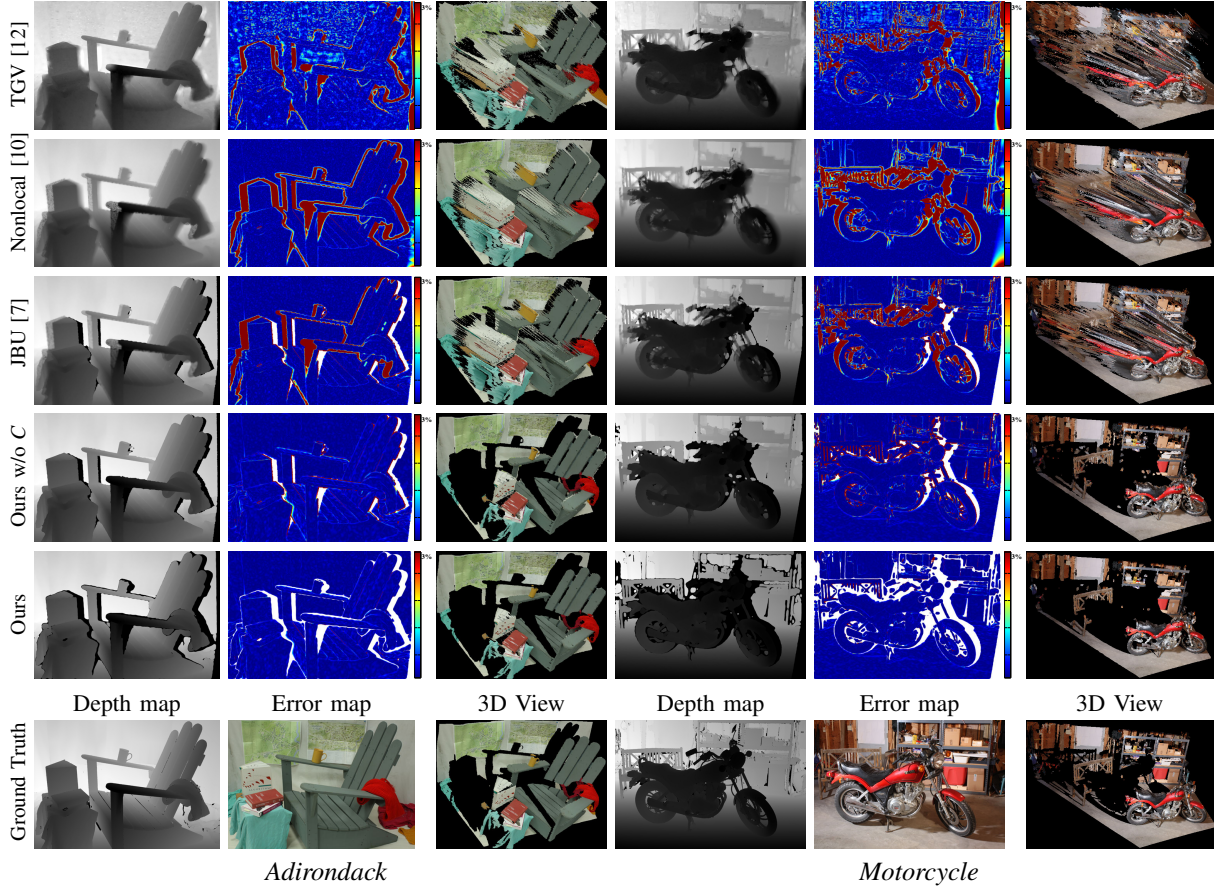

Fig. 7: Examples of upsampling results described in Section V-A. "Ours w/o $C$" denotes our upsampling method without our confidence map. The error maps depict a relative depth error ranging from 0 to the 3% of the maximum depth. The white pixels in the error maps are excluded when computing the results in Table I. We used $\sigma_s = 20$pixels, $\sigma_i = 20$pixels, and $\sigma_r = 30$mm for the experiment.

To tackle the problem, we designed a new testbed based on the Middlebury stereo 2014 datasets [20]. Each dataset in [20] consists of high-resolution stereo images, ground-truth depth maps estimated using a structured lighting system, and calibration parameters. In our testbed, we simulated the depth and image measurements as follows. First, we sampled the ground-truth depth pixels on the left view by every 4 pixels in column and 12 pixels in row to simulate asymmetric density of depth measurements of our system. We added additive Gaussian noise ($\sigma = 10$mm) on the samples. Then, the noisy depth samples were projected onto the right view. The aligned depth points cover about 2% of the image resolution. In total, we generated 23 test datasets from all the Middlebury stereo 2014 datasets with ground-truth. Though we used the ground-truth calibration parameters, the dataset generated from our testbed naturally exhibits the flying points, flipping points, and dis-occlusion problems similar to the dataset captured from our real system.

For a quantitative comparison, we use a robust accuracy measure, "A$N$" as used in [21]; "A$N$" denotes the depth error at the $N$-th percentile after sorting the errors from low to high. We use the hole-filled results for the global methods [12], [10], while for the local methods, JBU [7] and ours, we exclude the mask regions that cannot compute results with local filters due to large hole or low confidence.

Table I shows the quantitative comparison results. Our method works consistently well in the configurations both of A80 and A95, while the performance of the other methods are exponentially degraded in A95. The upsampling results and error maps are also shown in Fig. 7. Compared to our method, the other methods have large error and suffer from severe artifacts at the depth boundary regions that are clearly shown on the 3D view in the figure.

The major benefit of our approach is a novel depth outliers rejection scheme that gives clear seed depth points. In addition, our scale-aware depth upsampling gives more tolerance on the noisy depth measurements in homogeneous surfaces. The remaining ambiguous depth pixels adhered to the boundary region of a large structure are effectively rejected by our confidence map. The 3D views in Fig. 7 also show that our results successfully preserves depth discontinuity and its fine structures. More quantitative and qualitative comparison results are presented in the supplemental material[1].

### B. DRC finals 2015

Our depth upsampling algorithm was utilized for the part of DRC tasks; object detection and pose estimation of real-world objects. In the tasks, it was important to achieve dense and accurate depth map to successfully recognize 3D pose of target objects for grasping. In practice, raw depth data from lidar was too sparse and noisy to estimate the accurate pose of an object. Our depth upsampling method generated a high-quality depth map that improves the success-rate of each tasks.

We developed the task-specific detection algorithms for the three target objects; valve, drill, and terrain. In the VALVE and DRILL cases, we initially set a region of interest for a target object, then our pose estimation algorithms estimate 3D pose of a target object by fitting predefined 3D templates to a depth map. In the TERRAIN case, each local plane is detected by progressively grouping surface normal directions, then the group labels are refined using Graph Cut [22].

We performed the tasks with several depth maps variants such as raw depth map, our upsampling result, and state-of-the-art upsampling results in [12], [10], [7]. Fig. 8 shows the qualitative comparison of the upsampled depth maps. Fig. 9 visualizes the results of object detection and pose estimation. Our algorithm estimates an accurate and outliers-free depth map with sharp depth boundaries. Our depth map results in accurate 3D template fitting while the other tested variants may fail to detect accurate poses of objects. Especially, in the DRILL example, pose estimation is challenging because of the lack of valid points and depth error existing at the object boundaries. Using our depth map, the desirable grasping direction is correctly determined. In the case of TERRAIN detection, accurate estimation of surface normal is important for humanoid robot treading. Our method results in the most accurate pose estimation without any missing cinder block.

## VI. CONCLUSIONS

We have presented a vision system including calibration method and a depth processing method that are specially designed for robust and reliable depth sensing. Our depth processing method explicitly handles noisy or unreliable depth observations, therefore it outputs high-fidelity dense depth map. Through the intensive evaluations, we verified that our method outperforms over the state-of-the-art methods and demonstrated that our method is especially suitable for robotics applications.

---

[1]The high-resolution version can be found in here.

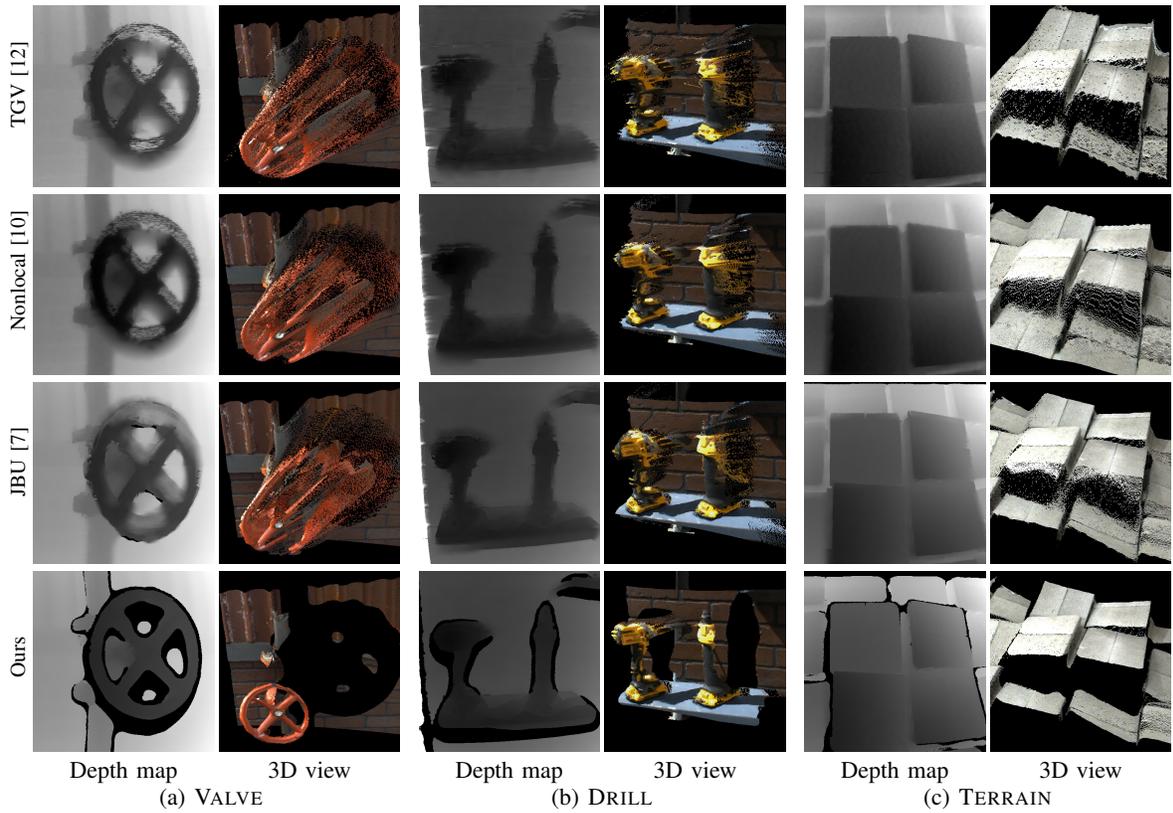

Fig. 8: Qualitative comparison of upsampled depth maps on the *DRC finals* 2015 datasets: VALVE, DRILL, and TERRAIN. The sparse depth observations acquired from lidar are propagated to generate dense depth maps. Note that clear depth boundaries observable in our results.

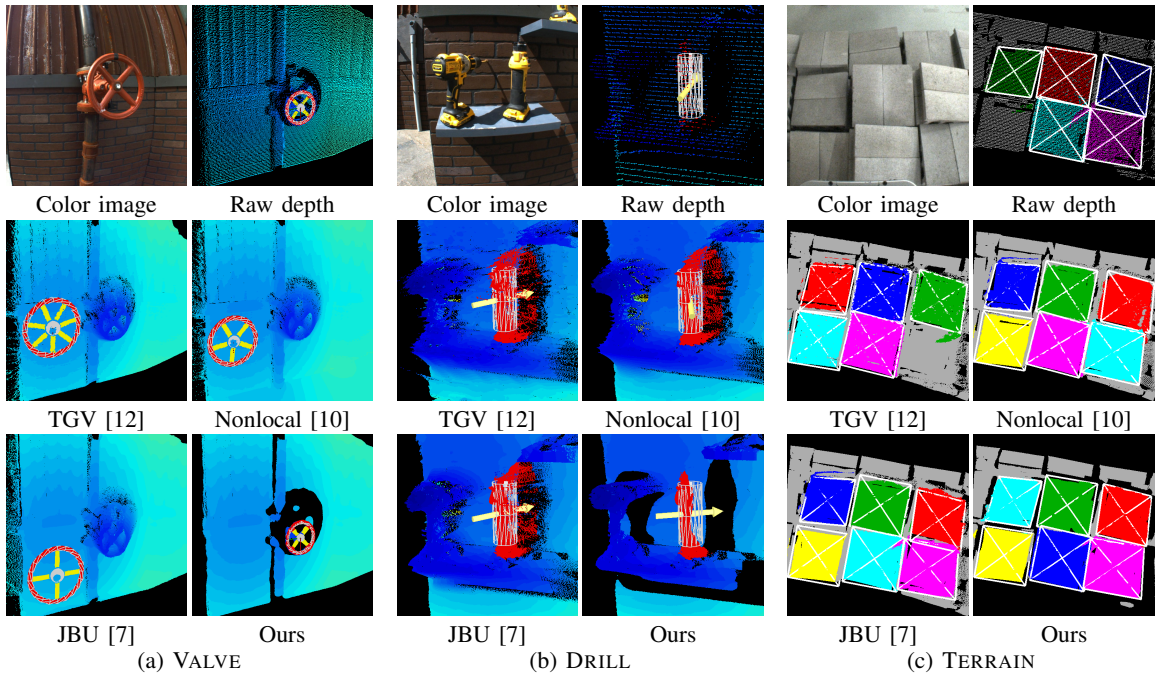

Fig. 9: Object detection and pose estimation using various depth maps. (a) VALVE. The 3D template is correctly aligned with our depth map. Note that our pose estimation approach evaluates several templates having different number of spokes and scales. (b) DRILL. The raw depth is too sparse, and it suffers from flying pixels. Although body part (white cylinder) are detected in every depth maps, the yellow arrow indicating grasping direction is correctly detected (corresponds to the dark part of the right drill in the color image) in our depth map. (c) TERRAIN. The colored points indicate the detected cinder blocks, and white rectangles denote the estimated pose of detected blocks.